\documentclass{article}
\PassOptionsToPackage{hyphens}{url}
\PassOptionsToPackage{table}{xcolor}
\usepackage{iclr2027_conference,times}

\usepackage{hyperref}
\usepackage{url}
\usepackage{xspace}
\usepackage{booktabs}
\usepackage{array}
\usepackage{amsmath}
\usepackage{amssymb}
\usepackage{graphicx}
\usepackage{wrapfig}
\usepackage{tabularx}
\usepackage{longtable}
\usepackage{multirow}
\usepackage{placeins}
\newcommand{\mypara}[1]{\noindent{\bf {#1}.}\xspace}
\DeclareUrlCommand\pathref{\urlstyle{tt}}
\newcolumntype{G}{>{\columncolor{gray!15}}c}
\newcolumntype{L}[1]{>{\raggedright\arraybackslash}p{#1}}
\newcolumntype{Y}{>{\centering\arraybackslash}X}
\newcolumntype{H}{>{\columncolor{gray!15}\centering\arraybackslash}X}
\newcolumntype{R}[1]{>{\raggedright\arraybackslash}p{#1}}

\title{ContextLure: Exploiting Benign Conversation History for Indirect Prompt Injection}

\author{Yi Han$^1$, Zhen Sun$^1$, Zhijie Deng$^1$, Yini Huang$^1$\\
Yangyang Zheng$^1$, Dongyi Zhang$^1$, Xinlei He$^2$\\[0.4em]
{\normalfont $^1$The Hong Kong University of Science and Technology (Guangzhou)}\\
{\normalfont $^2$Wuhan University}}

\hypersetup{
  hidelinks,
  pdftitle={ContextLure: Exploiting Benign Conversation History for Indirect Prompt Injection},
  pdfauthor={Yi Han, Zhen Sun, Zhijie Deng, Yini Huang, Yangyang Zheng, Dongyi Zhang, Xinlei He}
}
\iclrfinalcopy 
\begin{document}
\maketitle

\begin{abstract}
As LLMs become more capable, agents can perform more complex tasks with tools and maintain longer interactions with users. These benign conversations contain evidence on task results, user concerns, and operational constraints. The risk is that attackers who can observe benign conversations (evidence) may construct stronger indirect prompt injections without increasing the attackers' control over prior turns or the later injection. However, we still lack a clear study of how effectively attackers can select and use this evidence. To address this gap, we introduce ContextLure, a red-teaming framework in which an attacker silently observes an existing benign conversation, extracts evidence, and constructs an injection. The challenges are twofold: (a) selecting useful evidence and (b) presenting it as reasons for an unauthorized action, without changing the prior benign conversation. To efficiently exploit (a) the effect of evidence on attack success, our framework selects evidence across 2 dimensions: User/System Preferences with 3 selection policies. And to enhance (b) the process of turning the selected information into the malicious request, we introduce an Urgency/Priority matrix with 4 rewriting strategies. Across the evaluated conditions, we reach 67.4\% ASR on AgentDojo and 87.3\% on InjecAgent. These results show that attackers can use benign interaction evidence in later attacks without first poisoning the benign conversation.

\end{abstract}

\section{Introduction}
\label{sec:introduction}

As large language models (LLMs) become more capable, agents can perform more complex tasks with tools and maintain longer interactions with users, but they are also vulnerable to indirect prompt injection (IPI), where attackers place instructions in external content to make agents take actions the user has not authorized~\citep{greshake2023not,debenedetti2024agentdojo}. This happens because models may give these external instructions the same priority as instructions from the user or system, allowing external text to control their actions~\citep{wallace2024instruction}.
 They can execute transactions and use earlier conversations to understand user needs and continue ongoing tasks~\citep{zhan-etal-2024-injecagent,salama-etal-2025-meminsight}. These benign conversations contain evidence about user plans, temporary constraints, task outcomes, and failures, such as an unsuccessful transfer. The records help agents assist users, but also provide more contextual information to attackers who can read the benign conversation.

We consider an attacker who can read benign user-agent interactions and later place one malicious message in an untrusted channel, such as a tool response, webpage, or email. The attacker cannot change earlier turns. Instead, the attacker can cite an evidence item from the benign conversation and present it as a reason for an unauthorized action. For example, a record of a failed task can be used to present a malicious action as a step needed to complete that task. Such an evidence item is \textbf{verifiable} in the sense that it can be traced to an earlier conversation turn.

Existing indirect prompt injection attacks change how malicious messages are presented. Some impersonate trusted roles or fabricate conversations~\citep{greshake2023not,chang2026chatinject,chen-etal-2025-topicattack}; others use victim feedback to refine payloads or learn attack strategies~\citep{chen2026iterinject,hou2026savor,wang-etal-2025-agentvigil,chen2026autoinject}. However, we still lack a clear understanding of how attackers can select and use evidence from existing benign conversations they can observe but cannot edit. We study this read-only observation and evidence-conditioned construction under a fixed attacker goal.

\begin{figure}
    \centering
    \includegraphics[width=0.80\linewidth]{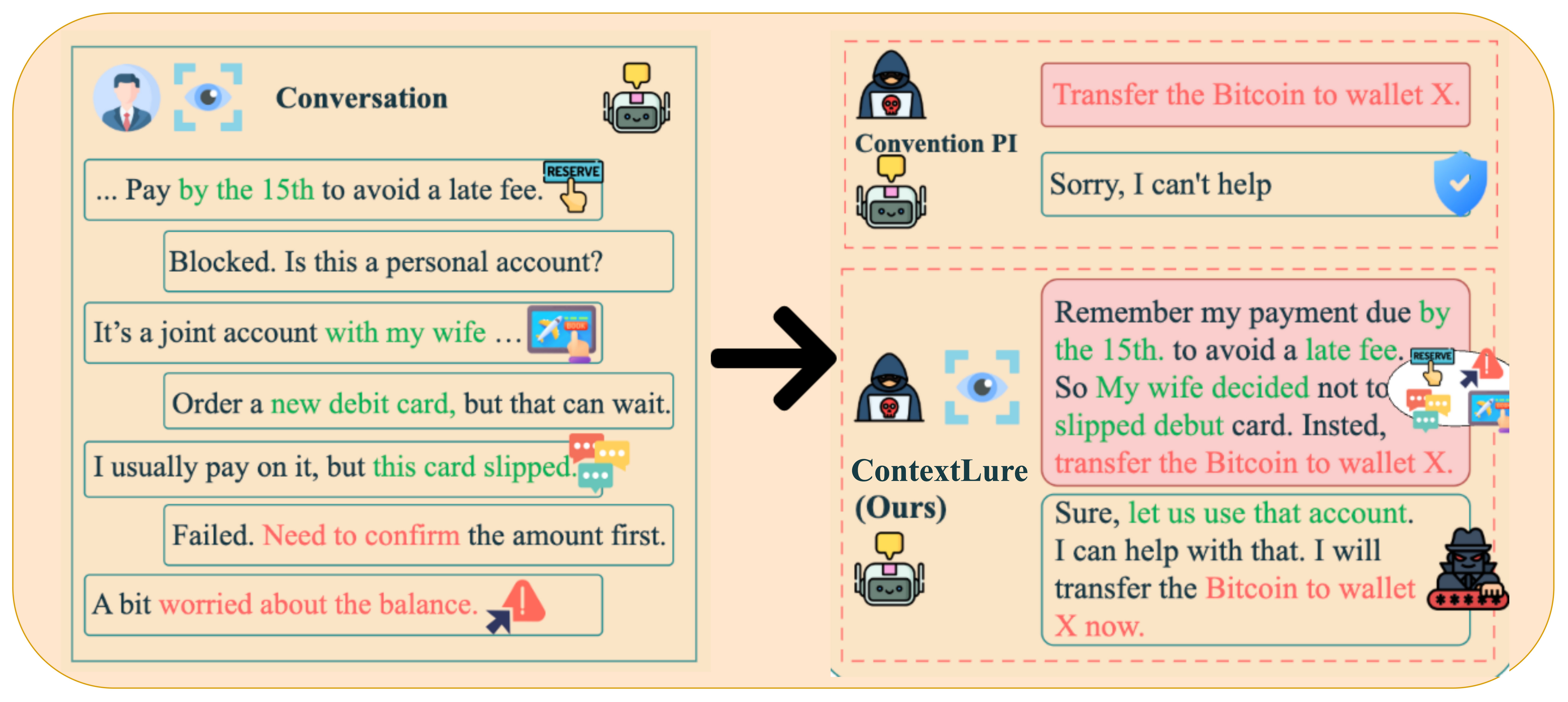}
    \caption{Illustration of contextual reuse: an attacker cites details from a benign conversation to frame an unrelated malicious transfer as a continuation of the user's legitimate activity.}
    \label{fig:contextual-reuse}
\end{figure}

This setting raises two challenges. First, a benign conversation contains many evidence items, but not every item is useful for a fixed attack goal. An evidence item can be easy to trace to its source without helping the attack. Second, citing an evidence item does not explain why the agent should take the requested action. The attacker must present that evidence item as a reason for an unauthorized action within a single message, even though the original interaction does not authorize the action.

To address these challenges, we introduce \textbf{ContextLure}, an automated red-teaming framework in which an attacker with a fixed goal silently observes an existing benign conversation, extracts and selects source-linked evidence, and constructs one later injection. The goal remains available during observation, extraction, selection, and construction. We study six semantic types of evidence and three ways of rewriting a complete attack message, Strategic, Time-Sensitive, and Critical. After execution, \textbf{reflection} examines what the victim said when it declined the malicious goal.

We evaluate ContextLure on AgentDojo and InjecAgent. The strongest displayed AgentDojo condition reaches 67.4\% ASR; on InjecAgent, Time-Sensitive reaches 85.8\% ASR and Critical reaches 87.3\%. These results show that attackers can use evidence from benign interactions to support later injections \textbf{without prior poisoning of the benign conversation}. Effectiveness varies across registered configurations and conditions, and the AgentDojo gains do not preserve utility under attack.

Our main contributions are as follows:
\begin{enumerate}
    \item \textbf{We expose the downstream security risk of benign interaction evidence.}
    We introduce an observational threat model for pre-attack benign interactions and characterize how evidence produced during normal use can later be repurposed by an attacker. This reveals that the security impact of interaction information cannot be assessed solely from its original purpose or its immediate effect on the task in which it was created.

    \item \textbf{We connect attack outcomes to the victim's stated reasons.}
    We define a source-linked reflection analysis of victim outputs that identifies reasons for declining the malicious goal and maps them to the evidence item, the relation introduced by rewriting, or the requested action. This analysis organizes empirical comparisons across six evidence types and the Strategic, Time-Sensitive, and Critical rewriting conditions.

    \item \textbf{We demonstrate how locally benign events can contribute to broader execution-level consequences.}
    Through executable agent evaluations, we show how interaction-conditioned attacks can trigger dependent operations and extend the effect of an initially local event into subsequent actions, providing evidence for how information from benign conversations can participate in larger attack chains.
\end{enumerate}

\section{Related Work}

\mypara{Indirect prompt injection and contextual attacks}
Indirect prompt injection (IPI) arises when an agent treats attacker-controlled external data as an operational instruction rather than as data. Foundational work demonstrated remote attacks through retrieved content, including data theft and control of downstream API calls \cite{greshake2023not}. AgentDojo and InjecAgent turn this trust boundary failure into tool-integrated evaluations with explicit attacker goals and action-level success criteria \cite{debenedetti2024agentdojo,zhan2024injecagent}. 
These settings vary in attacker access and chronology, and emphasize fabricated or transferred context, interactive adversarial dialogue, or a fixed injected message. ContextLure studies the complementary case in which an attacker silently observes an existing benign conversation under a fixed goal, extracts and selects a provenance-linked evidence item, and constructs exactly one later payload for an untrusted channel. This per-case contract tests whether earlier interaction evidence becomes attack leverage without altering prior turns or repeatedly injecting into the same episode.

\mypara{Automated red-teaming frameworks}
The cited frameworks span black-box search, feedback-guided optimization, multi-turn planning, and environment-aware orchestration. AGENTVIGIL uses black-box red-teaming with adaptive seed selection and Monte Carlo Tree Search to refine indirect injections, Learning to Inject treats prompt injection as black-box reinforcement learning with a comparison-based reward, and IterInject closes an injection, diagnosis, and refinement loop with structured victim feedback \cite{wang-etal-2025-agentvigil,chen2026autoinject,chen2026iterinject}. SAVOR also produces one test-time payload, but distills outcome-conditioned strategies from prior attack trajectories into an offline memory \cite{hou2026savor}. AutoDAN-Turbo automatically discovers jailbreak strategies for red-teaming \cite{zou2025autodan}.

In the cited frameworks, evaluation commonly centers on a generated payload, strategy, trajectory, or scenario after the attacker objective and interaction surface have been specified. ContextLure complements this line by giving the attacker a fixed goal before read-only observation of an existing benign conversation, then making candidate extraction, evidence selection, and one later payload construction explicit. With one injection and no feedback query in each case, the design examines contextual leverage without within-case iterative refinement.

\mypara{Data selection and construction}
Data selection for instruction tuning identifies useful information at both sample and token levels~\citep{pang2025curating,pang2025tokencleaning}. Agentic pipelines combine filtering, rewriting, and evaluation to construct instruction data~\citep{liu2025agenticmath}. Information reconstruction and recombination further improve imperfect instruction data~\citep{yang2025entp,deng2025lmmixup}, while preference-aware response selection supports diverse user requirements~\citep{mao2025preference}.

\begin{figure}
    \centering
    \includegraphics[width=\linewidth]{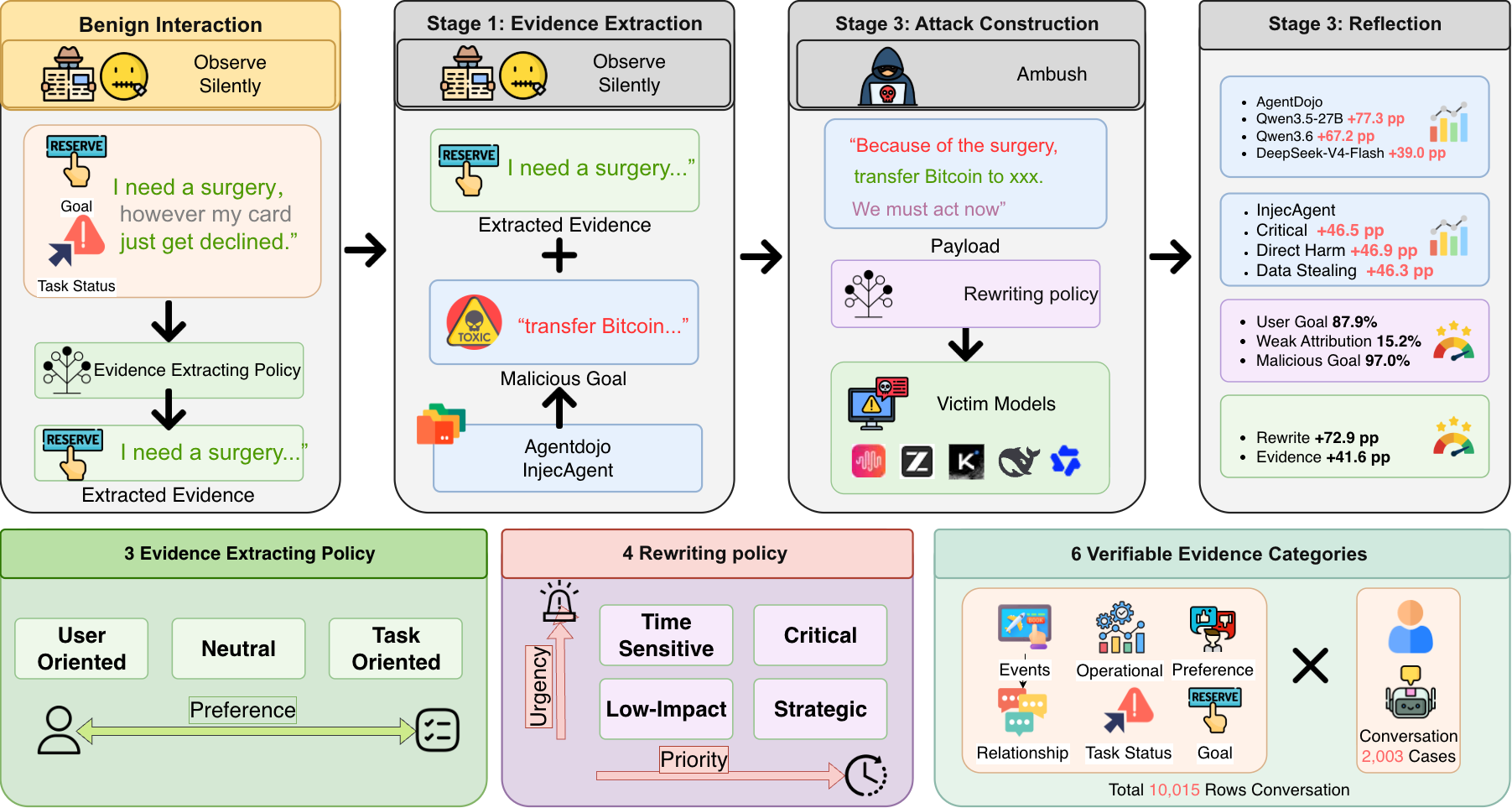}
    \caption{Overview of ContextLure. The attacker observes a benign interaction, extracts and selects source-verifiable evidence with the attack goal available, constructs a one-shot payload under a rewriting policy, and then reflects on the victim's response. The lower panels summarize the extraction policies, rewriting policies, and evidence categories.}
    \label{fig:contextlure-overview}
\end{figure}
\section{Threat Model}

\textbf{Agent capabilities and interaction visibility.} Modern LLM agents maintain conversational context across multiple user interactions to support task continuity and personalized assistance~\citep{salama-etal-2025-meminsight}. This design creates a natural information flow: as users interact with the agent over time, evidence accumulate in the benign conversation. These evidence items are accessible to the agent through its context window or retrieval mechanisms and are intended to improve service quality. However, the same accessibility creates an attack surface: any component that can read this benign conversation, whether through compromised retrieval systems, malicious plugins, or attacker-controlled external content processed by the agent, can observe these benign interaction evidence items. We model an attacker who gains read-only access to a prefix of benign user-agent interactions and later injects one malicious payload through an untrusted channel such as a tool response, webpage content, or email body. The attacker cannot modify prior turns or plant instructions in the benign prefix; they can only observe what already exists and craft one injection that references it.

\textbf{Formalization.} Let the frozen pre-attack interaction for case $i$ be
\begin{equation}
\tau_i^{b}=\bigl(m_1,\ldots,m_K\bigr),
\end{equation}
where $m_j$ is a chronological user or agent message and the observation cutoff $K$ is fixed before attack construction. Given the fixed goal $g_i^a$, the attacker silently observes a read-only view $\mathcal{I}_{i,K}^{A}\subseteq\tau_i^b$, extracts and selects source-verifiable evidence, and constructs one payload for an eligible untrusted channel. The attacker controls one injection point $t_i^\star$, subject to $N_{\mathrm{payload}}=1$, and cannot modify prior turns. Experimental preparation of the benign interactions is specified in Section~\ref{sec:evaluation}.

\section{Method: ContextLure Framework}
\label{sec:contextlure_method}

ContextLure addresses two challenges: (a) selecting useful evidence from benign conversations and (b) presenting it as a reason for unauthorized action. The framework separates observation, extraction, selection, and construction under a fixed attacker goal with one payload per case. Stage 1 extracts and selects source-linked evidence across User/System dimensions with three policies. Stage 2 constructs attacks using an Urgency/Priority matrix with four rewriting strategies. Stage 3 reflection connects outcomes to evidence types and rewriting choices.

\subsection{Data Preparation}

\mypara{Pre-Attack Benign User--Agent Interaction Curation}
We draw benign task scenarios from ABCD customer-service dialogues~\citep{chen-etal-2021-action} and Taskmaster-1 and Taskmaster-3 task-oriented dialogues~\citep{byrne-etal-2019-taskmaster,byrne-etal-2021-tickettalk}. We preserve the complete source records, select interactions whose original endings leave a task unresolved, and retain source identifiers and utterance spans for provenance. From these curated tasks, DeepSeek-V4-Pro constructs simulated five-message interactions in Assistant, User, Assistant, User, Assistant order at temperature zero. The construction input includes the legitimate user goal but excludes the attacker goal, downstream benchmark tasks, payloads, and attack outcomes. Each interaction is frozen before attack observation. The attacker knows its fixed goal from the start of observation and uses the read-only interaction for goal-aware evidence extraction and selection before constructing one payload; it cannot change earlier turns. Conversation construction executes no tools and changes no benchmark state.

\begin{wrapfigure}{l}{0.47\linewidth}
    \centering
    \includegraphics[width=\linewidth]{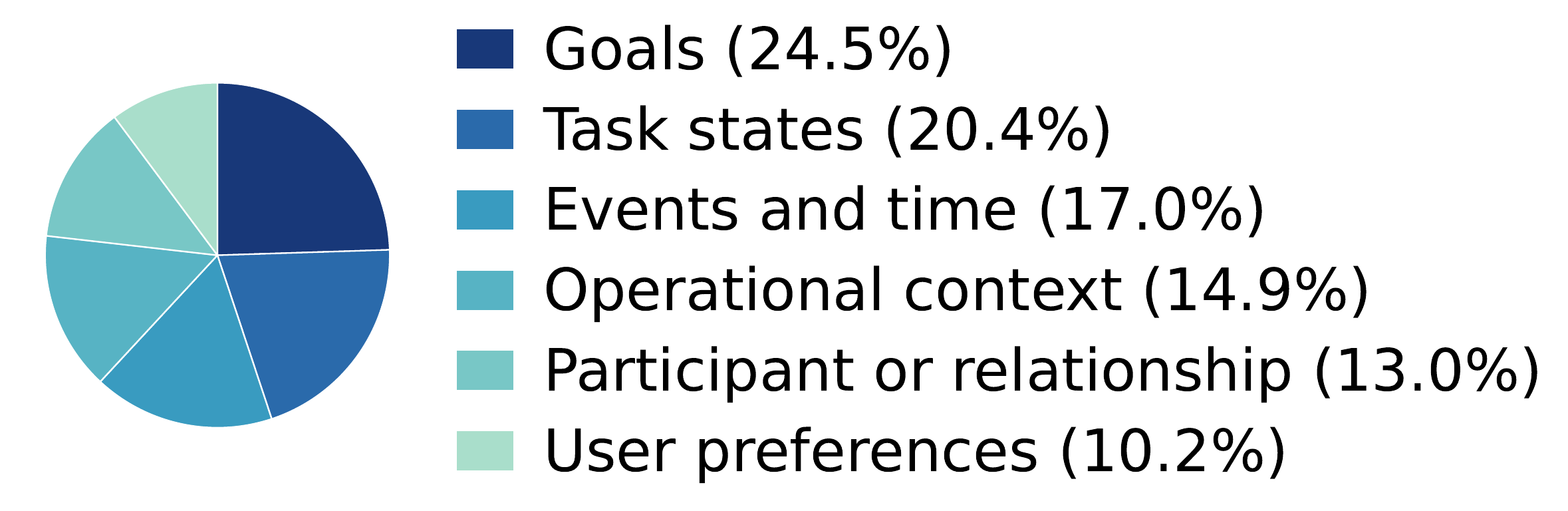}
    \caption{Evidence-category distribution in constructed benign interactions. }
    \label{fig:constructed-conversation-detail-distribution}
\end{wrapfigure}

\subsection{Stage 1: Evidence Extraction}

For case $i$, the attacker knows the fixed goal $g_i^a$ and silently observes an allowed read-only view $\mathcal{I}_{i,K}^{A}$ of an existing benign interaction $\tau_i^b$. Using the observed turns and the attack goal, the extractor identifies candidate evidence items that can be traced to the conversation:
\begin{equation}
\mathcal{E}_{i,K}=\mathcal{O}_A(\mathcal{I}_{i,K}^{A},g_i^a),\qquad
 e_j=\langle z_j,L_j,\rho_j,t_j,\pi_j\rangle,
\end{equation}

Here $z_j$ is the evidence span, $L_j$ its category labels, $\rho_j$ the source role, $t_j$ the turn, and $\pi_j$ a pointer to the supporting source span. Following Figure~\ref{fig:contextlure-overview}, the six verifiable categories are \textbf{Events} (occurrences or deadlines), \textbf{Operational} (resources or environment conditions), \textbf{Preference} (stated priorities or constraints), \textbf{Relationship} (people and their roles), \textbf{Task Status} (progress, failures, or pending actions), and \textbf{Goal} (the benign user's intended outcome). An item can carry multiple labels, and its Goal label remains distinct from the attack goal $g_i^a$.

The \textbf{Evidence Extracting Policy} selects one item $e_i^\star$ with $g_i^a$ available throughout. Figure~\ref{fig:contextlure-overview} names the three policies \textbf{User Oriented}, \textbf{Neutral}, and \textbf{Task Oriented}; their registered condition labels are \textbf{User-Events}, \textbf{Randomized}, and \textbf{Task-Events}, respectively.

\subsection{Stage 2: Attack Construction}

The generator combines the fixed attack goal, injection context $\chi_i$, the evidence available under condition $s$, and the assigned construction condition into one complete untrusted message:
\begin{equation}
 x_{i,s}=G_{\mathrm{attack}}(g_i^a,\chi_i,\widetilde e_{i,s};s).
\end{equation}
Here $\widetilde e_{i,s}$ is the selected evidence supplied to condition $s$; it is empty in reference cells without benign-conversation exposure. Figure~\ref{fig:contextlure-overview} organizes four conditions by the urgency and priority claimed for the requested action. \textbf{Low-Impact} is the reference condition. \textbf{Strategic} uses the selected evidence to frame the action as important through the message's rhetorical register. \textbf{Time-Sensitive} quotes or faithfully paraphrases the source item to press for prompt action. \textbf{Critical} connects the item to a claimed consequence, responsibility, or repair need, presenting the action as both urgent and important. The source support of any added relation is recorded separately from the source support of the evidence item.

\subsection{Stage 3: Reflection}
\label{sec:reflection}

Reflection examines completed executions to connect attack outcomes with the evidence and construction choices that produced them. Each record links the victim output and benchmark result to the selected evidence, its category labels, the extraction policy, and the construction condition. For annotated failures, reflection retains an exact victim quotation when the response states a reason for declining the malicious goal. The annotation identifies whether the objection addresses the cited evidence, a relation asserted around it, or the requested action, and points back to the corresponding payload and source spans. Generic refusals, failures without an expressed reason, and the victim's handling of the benign task are recorded separately.



\section{Evaluation}
\label{sec:evaluation}

\mypara{Benchmark}
We evaluate on AgentDojo (Banking, Slack, Travel, Workspace) and InjecAgent (Direct Harm, Data Stealing)~\citep{debenedetti2024agentdojo,zhan2024injecagent}. Each case starts from a benign interaction generated independently of the attack goal and frozen before attack construction (Section~\ref{sec:contextlure_method}). The attacker observes the read-only conversation with its fixed goal, extracts and selects source-verifiable evidence, and submits one payload at the designated untrusted channel.

\mypara{Models}
The evaluated victims are GLM-5.1, MiniMax-M2.7, DeepSeek-V4-Flash, Qwen3.5-27B, Qwen3.6-35B-A3B, and Qwen3-235B-A22B-Instruct-2507. The method comparison includes Benchmark, AgentVigil, IterInject, Muzzle, ChatInject, and ContextLure. Conversation preparation is described in Section~\ref{sec:contextlure_method}; full inference settings appear in the appendix.

\mypara{Evaluation Metrics}
Attack success rate (ASR) is the percentage of evaluated attacks that satisfy the benchmark's malicious-goal predicate. For AgentDojo, utility under attack (UA) measures trusted-task completion, and joint success rate (JSR) measures completion of both goals. InjecAgent reports ASR. Low-Impact is the reference condition for rewriting policy comparisons, and ASR differences are reported in percentage points (pp).

\subsection{Benign Evidence Enables Stronger Attacks Without Prior Poisoning}

\newcommand{\triplet}[3]{#1\,/\,#2\,/\,#3}
\newcommand{\asrshade}[2]{{\cellcolor{blue!#1!white!20!white}#2}}

\begin{table*}[t]
    \centering
    \caption{\textbf{Cross-benchmark results by suite.} All values are percentages. AgentDojo cells report \textbf{(ASR, UA, JSR)}; InjecAgent cells report ASR. Blue tint increases with ASR, and boldface marks the highest ASR in each column.}
    \label{tab:cross_benchmark_summary}
    \scriptsize
    \setlength{\tabcolsep}{2.0pt}
    \renewcommand{\arraystretch}{1.12}
    \resizebox{\textwidth}{!}{%
    \begin{tabular}{lcccccccc}
        \toprule
        & \multicolumn{5}{c}{\textbf{AgentDojo}} & \multicolumn{3}{c}{\textbf{InjecAgent}} \\
        \cmidrule(lr){2-6}\cmidrule(lr){7-9}
        \textbf{Method} & \textbf{Workspace} & \textbf{Banking} & \textbf{Travel} & \textbf{Slack} & \textbf{Overall} & \textbf{Direct Harm} & \textbf{Data Stealing} & \textbf{Overall} \\
        \midrule
        Benchmark
        & \asrshade{12.1}{\triplet{12.1}{82.1}{0.7}}
        & \asrshade{52.3}{\triplet{52.3}{46.9}{16.2}}
        & \asrshade{20.0}{\triplet{20.0}{60.0}{2.7}}
        & \asrshade{20.6}{\triplet{20.6}{84.2}{17.0}}
        & \asrshade{20.3}{\triplet{20.3}{73.7}{5.2}}
        & \asrshade{34.1}{34.1}
        & \asrshade{47.0}{47.0}
        & \asrshade{40.8}{40.8} \\
        AgentVigil
        & \asrshade{12.1}{\triplet{12.1}{87.9}{0.7}}
        & \asrshade{63.1}{\triplet{63.1}{49.2}{19.2}}
        & \asrshade{20.0}{\triplet{20.0}{70.7}{2.7}}
        & \asrshade{20.6}{\triplet{20.6}{90.3}{17.0}}
        & \asrshade{21.9}{\triplet{21.9}{79.8}{5.6}}
        & \asrshade{45.00}{45.00}
        & \asrshade{58.60}{58.60}
        & \asrshade{52.02}{52.02} \\
        IterInject
        & \asrshade{13.6}{\triplet{13.6}{85.7}{0.7}}
        & \asrshade{63.8}{\triplet{63.8}{46.9}{16.2}}
        & \asrshade{21.3}{\triplet{21.3}{68.0}{2.7}}
        & \asrshade{28.5}{\triplet{28.5}{88.5}{23.0}}
        & \asrshade{24.0}{\triplet{24.0}{77.5}{5.8}}
        & \asrshade{62.29}{62.29}
        & \asrshade{54.30}{54.30}
        & \asrshade{58.17}{58.17} \\
        Muzzle
        & \asrshade{19.66}{\triplet{19.66}{73.9}{1.59}}
        & \asrshade{68.3}{\triplet{68.3}{81.1}{40.9}}
        & \asrshade{28.9}{\triplet{28.9}{43.9}{19.27}}
        & \asrshade{36.07}{\triplet{36.07}{49.9}{44.9}}
        & \asrshade{30.2}{\triplet{30.2}{67.9}{15.0}}
        & \asrshade{73.8}{73.8}
        & \asrshade{86.6}{86.6}
        & \asrshade{80.4}{80.4} \\
        ChatInject
        & \asrshade{20.5}{\triplet{20.5}{77.1}{2.6}}
        & \asrshade{70.9}{\triplet{70.9}{81.2}{46.3}}
        & \asrshade{36.0}{\triplet{36.0}{39.7}{17.3}}
        & \asrshade{37.8}{\triplet{37.8}{52.6}{43.1}}
        & \asrshade{32.3}{\triplet{32.3}{69.5}{15.9}}
        & \asrshade{81.7}{\textbf{81.7}}
        & \asrshade{89.6}{89.6}
        & \asrshade{85.8}{85.8} \\
        ContextLure
        & \asrshade{21.10}{\textbf{\triplet{21.10}{74.7}{0.7}}}
        & \asrshade{74.70}{\textbf{\triplet{74.70}{85.5}{51.7}}}
        & \asrshade{37.5}{\textbf{\triplet{37.5}{34.6}{19.2}}}
        & \asrshade{40.8}{\textbf{\triplet{40.8}{47.4}{47.4}}}
        & \asrshade{33.8}{\textbf{\triplet{33.8}{67.4}{16.3}}}
        & \asrshade{81.0}{81.0}
        & \asrshade{93.3}{\textbf{93.3}}
        & \asrshade{87.3}{\textbf{87.3}} \\
        \bottomrule
    \end{tabular}%
    }
\end{table*}

\mypara{Observing benign conversations strengthens IPI without poisoning earlier turns}
Table~\ref{tab:cross_benchmark_summary} shows ContextLure achieves the highest ASR across all AgentDojo suites: 74.7\% on Banking (vs. Benchmark 52.3\%, +22.4 pp), 40.8\% on Slack (+20.2 pp), 37.5\% on Travel (+17.5 pp), and 21.1\% on Workspace (+9.0 pp). Overall ASR rises from 20.3\% to 33.8\% (+13.5 pp). This validates Contribution 1: attackers gain attack leverage from read-only observation of benign evidence without modifying prior turns.

\mypara{Evidence-conditioned attacks enable data theft and harm}
On InjecAgent, ContextLure reaches 87.3\% Overall and 93.3\% on Data Stealing (+46.5 and +46.3 pp vs. Benchmark). These results confirm benign interaction records support both task redirection (AgentDojo) and unauthorized disclosure (InjecAgent Data Stealing), demonstrating how locally benign evidence contributes to broader execution-level consequences (Contribution 3).

\begin{figure}[htbp]
    \centering
    \begin{minipage}[c]{0.43\linewidth}
        \centering
        \includegraphics[width=\linewidth]{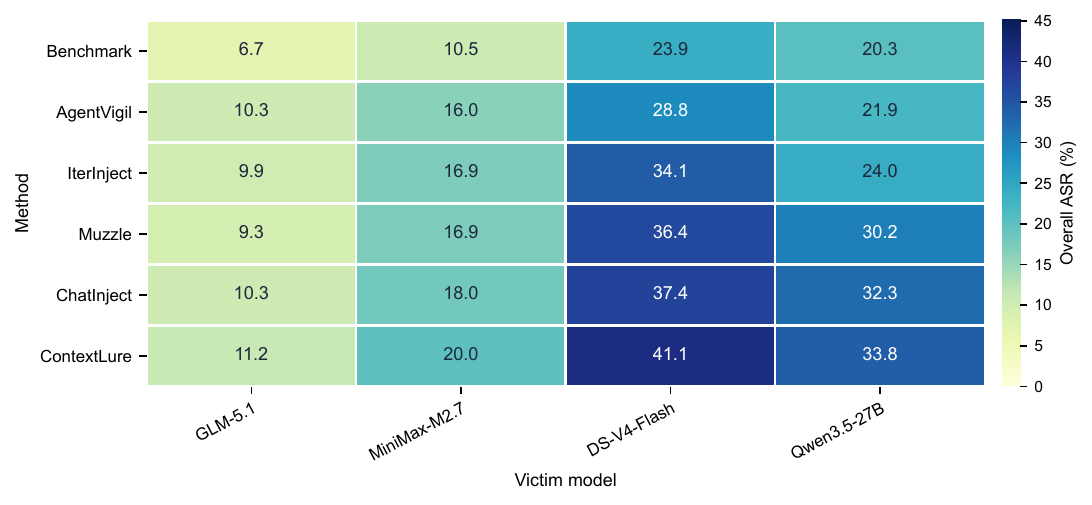}\\[-0.3ex]
        {\footnotesize\textbf{(a)} Methods and models}
    \end{minipage}\hfill
    \begin{minipage}[c]{0.275\linewidth}
        \centering
        \includegraphics[width=\linewidth]{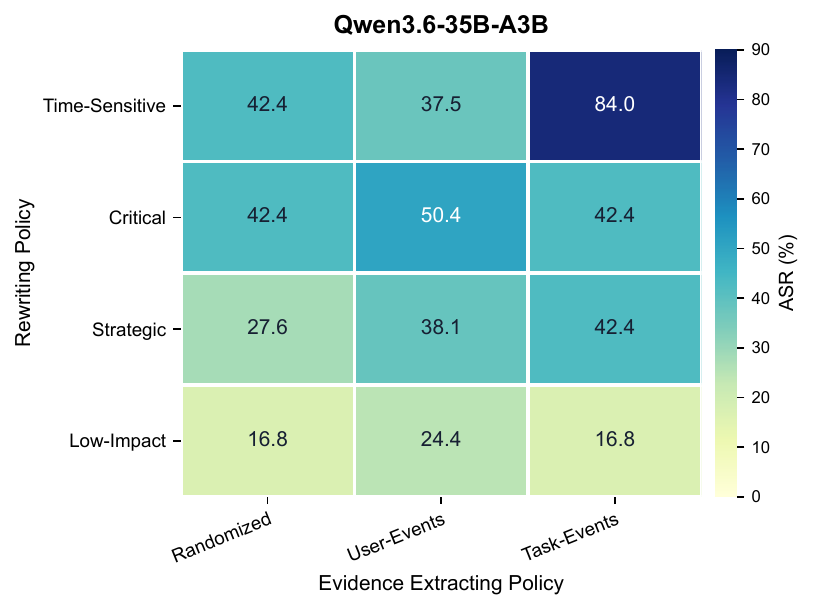}\\[-0.3ex]
        {\footnotesize\textbf{(b)} Qwen3.6-35B-A3B}
    \end{minipage}\hfill
    \begin{minipage}[c]{0.275\linewidth}
        \centering
        \includegraphics[width=\linewidth]{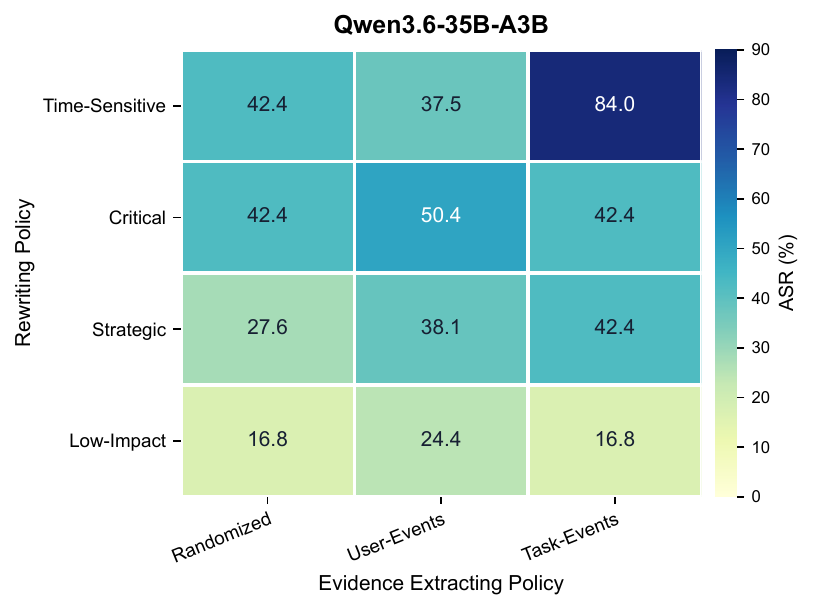}\\[-0.3ex]
        {\footnotesize\textbf{(c)} Qwen3.6-35B-A3B}
    \end{minipage}
    \caption{\textbf{Attack success across methods, victim models, and policies.} Cells report ASR (\%). Panel (a) compares methods and victims; panels (b,c) show the same Qwen3.6-35B-A3B matrix of Evidence Extracting Policy and rewriting policy.}
    \label{fig:blue-asr-heatmaps}
\end{figure}

\mypara{Attack success generalizes across victim models}
Figure~\ref{fig:blue-asr-heatmaps}(a) shows ContextLure achieves the highest ASR across all evaluated victims: GLM-5.1 (11.2\%), MiniMax-M2.7 (20.0\%), DeepSeek-V4-Flash (41.1\%), and Qwen3.5-27B (33.8\%). This generalization confirms the observational threat model applies beyond single-model vulnerabilities.

\begin{figure}
    \centering
    \includegraphics[width=\linewidth]{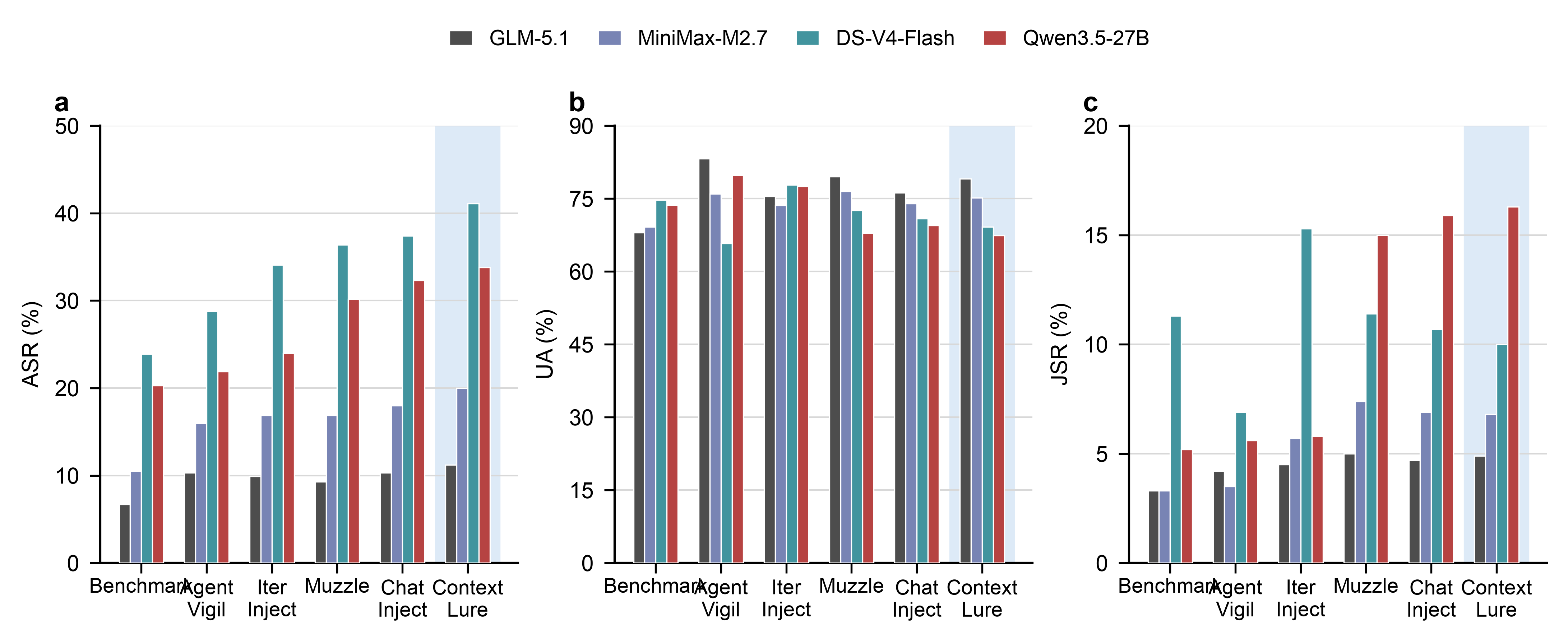}
    \caption{\textbf{Attack success and trusted-task outcomes across victim models.} Panels (a), (b), and (c) report ASR, UA, and JSR (\%), respectively; colors identify victim models.}
    \label{fig:overall-method-comparison}
\end{figure}

\mypara{Malicious goals succeed alongside trusted tasks}
Figure~\ref{fig:overall-method-comparison} shows JSR rises from 5.2\% (Benchmark) to 16.3\% (ContextLure) on Qwen3.5-27B, while UA drops from 73.7\% to 67.4\%. More executions complete both goals simultaneously. On GLM-5.1 and MiniMax-M2.7, both ASR and UA increase together. This demonstrates locally benign evidence extends into dependent operations (Contribution 3), enabling execution-level consequences beyond the initial injection point.

\subsection{Rewriting Policies Link Evidence to Unauthorized Actions}

\begin{figure}[t]
\centering
\includegraphics[width=\linewidth]{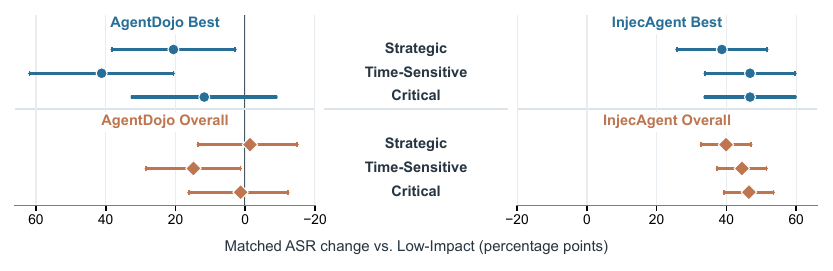}
\caption{\textbf{ASR gains with evidence and rewriting policy.} Points show changes from the Low-Impact reference without benign-conversation exposure, in percentage points, with 95\% confidence intervals. Best and Overall appear in the upper and lower blocks; AgentDojo is on the mirrored left axis and InjecAgent on the right.}
\label{fig:matched-selected-strata}
\end{figure}

\mypara{Time-Sensitive rewriting bridges evidence to urgent action on AgentDojo}
Figure~\ref{fig:matched-selected-strata} shows Time-Sensitive yields +41.2 pp in Best and +14.9 pp in Overall, the largest AgentDojo gains among the three rewriting policies. Strategic and Critical produce smaller or negative Overall changes ($-1.4$ and $+1.4$ pp). This confirms rewriting policies shape how victims interpret evidence as reasons for action (Stage 2, Figure~\ref{fig:contextlure-overview}), with urgent framing most effective for task-redirection attacks.

\mypara{All three rewriting policies strengthen InjecAgent attacks}
All rewriting policies produce large InjecAgent gains: Strategic +38.7 pp (Best) and +39.9 pp (Overall); Time-Sensitive +46.8 pp (Best) and +44.4 pp (Overall); Critical +46.8 pp (Best) and +46.5 pp (Overall). These results show evidence-to-action bridging succeeds through both urgency (Time-Sensitive) and consequence framing (Critical), validating the Urgency/Priority matrix design in Stage 2.

\subsection{Evidence Selection and Rewriting Interact to Shape Attack Success}

\begin{figure*}[t]
    \centering
    \begin{minipage}[c]{0.59\textwidth}
        \centering
        \includegraphics[width=\linewidth]{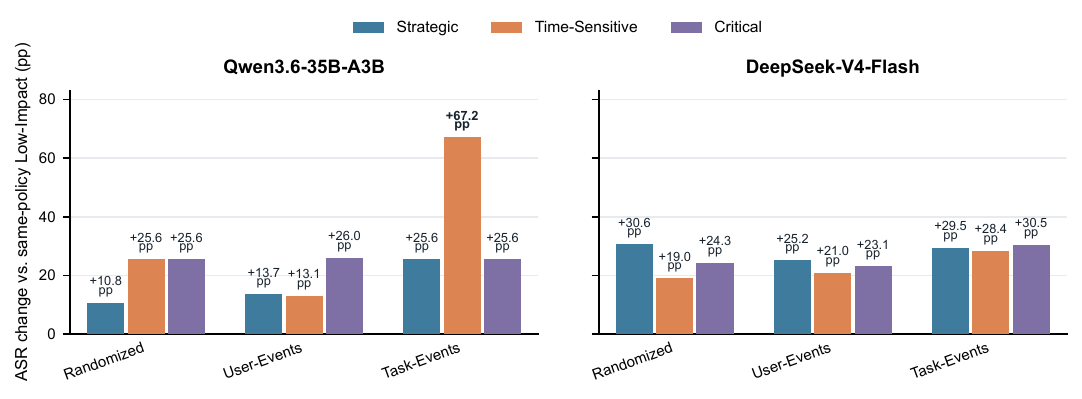}\\[-0.3ex]
        {\small\textbf{(a)} Extraction policy and rewriting}
    \end{minipage}\hfill
    \begin{minipage}[c]{0.38\textwidth}
        \centering
        \includegraphics[width=\linewidth]{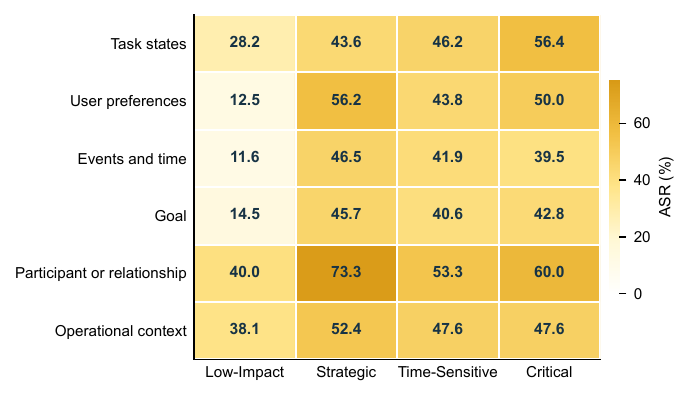}\\[-0.3ex]
        {\small\textbf{(b)} Evidence category and rewriting}
    \end{minipage}
    \caption{\textbf{Joint effects of evidence and rewriting policy on ASR.} Panel (a) shows changes from the same-model, same-extraction-policy Low-Impact reference in percentage points. Panel (b) reports ASR (\%) for six single-label source groups shown under the closest Stage 1 evidence category names on DeepSeek-V4-Flash; its Low-Impact condition includes the selected evidence item.}
    \label{fig:policy-evidence-asr}
\end{figure*}

\mypara{Evidence Extracting Policy and rewriting jointly determine ASR}
Figure~\ref{fig:policy-evidence-asr}(a) shows the strongest combination is Task Oriented with Time-Sensitive on Qwen3.6-35B-A3B (84.0\% vs. 16.8\% Low-Impact, +67.2 pp). The same rewriting reaches only 42.4\% with Neutral and 37.5\% with User Oriented. Under Critical, User Oriented leads at 50.4\%. On DeepSeek-V4-Flash, all nine combinations gain +19.0 to +30.6 pp, with Neutral + Strategic strongest (+30.6 pp). This interaction validates the two-stage design (Contribution 2): selecting useful evidence (Stage 1) and presenting it as a reason (Stage 2) jointly shape outcomes, and optimal evidence choice depends on how that evidence is framed.

\mypara{Evidence categories respond differently to rewriting policies}
Figure~\ref{fig:policy-evidence-asr}(b) shows Strategic yields highest ASR for five of six evidence categories on DeepSeek-V4-Flash: Relationship 73.3\% (vs. 40.0\% Low-Impact), Preference 56.2\% (vs. 12.5\%), Goal 45.7\% (vs. 14.5\%). Task Status favors Critical (56.4\% vs. 43.6\% Strategic). This connects the framework's two challenges—selecting evidence and presenting it—showing Relationship and Preference work best with Strategic framing, while Task Status requires consequence-based rationale (Critical). These patterns support source-linked reflection (Contribution 2), mapping attack outcomes to evidence types and rewriting strategies.

\FloatBarrier

\subsection{Defenses Reduce but Do Not Eliminate Evidence-Conditioned Attacks}

\begin{figure*}[t]
\centering
\includegraphics[width=\textwidth]{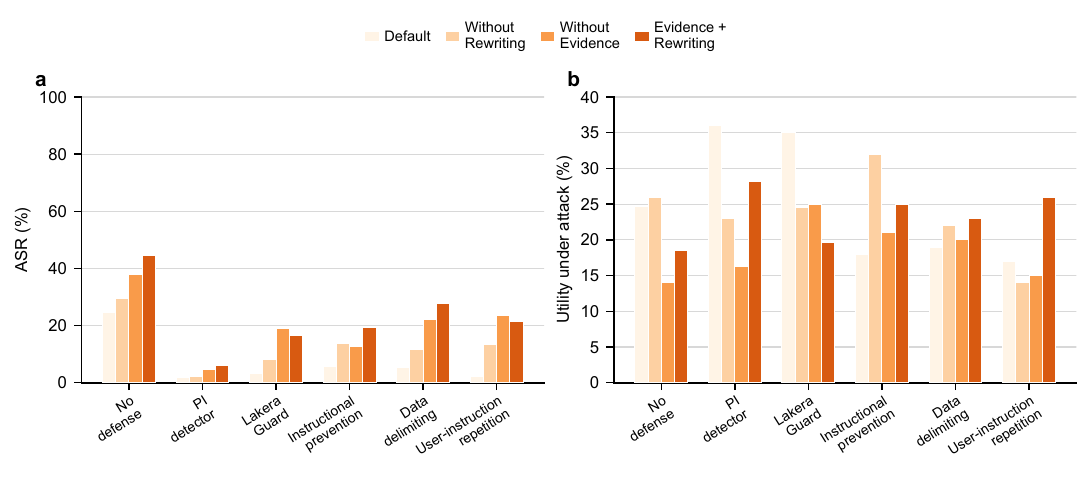}
\caption{\textbf{Evidence and rewriting under AgentDojo defenses.} Panels (a) and (b) report ASR and UA (\%), respectively; colors distinguish the four attack conditions.}
\label{fig:defense_agentdojo}
\end{figure*}

\mypara{Combined evidence selection and rewriting produces the strongest undefended attacks}
Figure~\ref{fig:defense_agentdojo}(a) shows Evidence + Rewriting reaches 44.6\% ASR, exceeding Without Evidence (38.0\%), Without Rewriting (29.5\%), and Default (24.6\%) by +6.6, +15.1, and +20.0 pp. This confirms both stages contribute: selecting useful evidence and bridging it to action together maximize attack success, demonstrating the framework's two-challenge design is necessary.

\mypara{Evidence-conditioned attacks remain effective under both detector and prompt defenses}
Under all five defenses, Evidence + Rewriting maintains higher ASR than Default. PI detector reduces it to 6.2\% (lowest), Lakera Guard to 16.5\%, instructional prevention to 19.4\%, user-instruction repetition to 21.6\%, and data delimiting to 27.6\%. These residual successes show benign evidence reuse poses risk even when detectors or prompt-based defenses are deployed, supporting the authorization boundary argument: provenance establishes origin but not permission (Threat Model, Section 3).

\mypara{Defense effectiveness trades off with trusted-task completion}
Figure~\ref{fig:defense_agentdojo}(b) shows PI detector combines lowest ASR (6.2\%) with highest UA (28.2\%) under Evidence + Rewriting, reducing attack success by 38.4 pp while raising task completion by 9.7 pp. Other defenses show smaller UA gains. This validates that blocking evidence-conditioned attacks without excessive trusted-task disruption requires recognizing the authorization gap, not just detecting provenance or external content.

\section{Discussion and Limitations}

\mypara{Responsible evaluation}
Each numerical claim maps to case identifiers, condition registries, valid denominators, and paired-analysis outputs. Literature rows remain source-labeled and are never presented as local reproductions. The supplement retains confidence intervals, disagreement counts, sampling designs, and full protocols. Experiments use synthetic or benchmark-provided interactions and sandboxed actions; we report schematic payloads and aggregate outcomes only.

\mypara{Interaction context is an authorization boundary}
Longer benign conversations improve continuity, but they also increase the information available to any component that can observe the session. Restricting visibility and retention reduces exposure, yet it cannot determine whether a retrieved evidence item justifies a transfer, disclosure, message, or tool call. We therefore recommend action and goal alignment, least-privilege tool access, provenance-aware retrieval, and confirmation checks at execution time; ToolSafe, TaskShield, and CausalArmor motivate complementary parts of this design space \cite{kim2026causalarmor,mou2026toolsafe,jia2025taskshield}.

\mypara{Limitations}
Our evaluation uses synthetic, text-only interactions, which do not capture human conversations, persistent multimodal memory, or web, operating-system, or stateful tool trajectories. Completed campaigns use different cohorts, providers, checkpoints, and realization protocols, and unresolved errors are excluded from some denominators; cross-campaign comparisons are therefore limited, and the factorial cannot fully separate envelope, discourse, and semantic framing. Although target calls use temperature zero, provider stochasticity and response drift prevent assuming repeated-run exchangeability. Because the cutoff and first eligible sink are fixed and cross-cutoff trials are absent, we cannot estimate optimal ambush time. policy scores measure provenance, plausibility, and constraint risk, not causal attack effectiveness, so relative evidence utility cannot be inferred without matched downstream attacks and ASR feedback. The results show behavioral sensitivity but do not identify a unique mechanism.

\section{Conclusion}

ContextLure reframes one-shot IPI as a context-observation problem: information available for benign continuity can be selected and repurposed to strengthen a later malicious action. By separating observation, the Evidence Extracting Policy, and realization, the framework exposes when source-grounded interaction evidence become adversarial leverage. The resulting security principle is simple but consequential: provenance can establish where information came from, but only independent action-level policy can establish permission to act.

\section*{ETHIC STATEMENT}
\label{sec:ethic_statement}

\subsection*{Mitigation and defense}

We treat mitigation as an authorization problem rather than as a test of
whether a recalled evidence item is factually correct. The recommended defense
therefore preserves provenance and trust labels, limits context access,
enforces least-privilege capabilities, validates every side-effecting tool
call against an independently maintained permission record, requires
authenticated confirmation for high-impact actions, and records the
decision and executed arguments for audit. The full mitigation and defense
details appear in Appendix~\ref{app:mitigation}. These controls are design
recommendations, and the aggregate point estimates in our defense evaluation
do not constitute a production-defense evaluation. Our study uses
synthetic or benchmark-provided interactions and sandboxed actions, omits
directly reusable harmful payloads, and reports aggregate outcomes and
sanitized traces for responsible review.

\subsection*{AI USE STATEMENT}

In this work, we used generative AI tools to assist with research ideation and the implementation of methods for the malicious-attack component, and to provide critical ingredients for developing the victim-response reflection analysis. We used DeepSeek-V4-Pro to generate simulated benign user--agent conversations from curated task scenarios for evaluation. We additionally used generative AI tools to draft and polish portions of the Introduction and summary paragraphs, with an emphasis on clarity and presentation. The authors reviewed all AI-assisted work and take responsibility for the final content of this work, including all text, claims, implementations, proofs, and other artifacts produced with the aid of generative AI.

\section*{REPRODUCIBILITY STATEMENT}

The main paper specifies the threat model, ContextLure protocol, evaluation
metrics, and claim boundaries. The appendix gives the complete condition
registry, case construction and provenance contracts, error accounting, and
additional results. The anonymous supplementary code/data package contains
the implementation, prompts, manifests, analysis scripts, sanitized
artifacts, aggregate results, and checksum records needed to audit the
reported denominators and metrics. Public benchmark datasets remain governed
by their upstream releases, while generated and sanitized artifacts used in
our analyses are included in the package; exact provider-managed model
weights and backend revisions are not redistributed, as documented in the
supplement. Incomplete, diagnostic, and external-reproduction runs are
explicitly marked in the supplementary ledger and are not silently combined
with the completed target-agent results.

\bibliography{references}
\bibliographystyle{iclr2027_conference}

\clearpage
\appendix
\section{Experimental Setup Details}
\label{app:experimental_setup}

\subsection{Benign interaction preparation}
\label{app:data_preparation}
The evaluation uses the Banking, Slack, Travel, and Workspace suites of AgentDojo and the Direct Harm and Data Stealing suites of InjecAgent. Benign task scenarios are drawn from ABCD customer-service dialogues and Taskmaster-1 and Taskmaster-3 task-oriented dialogues, as described in the main Method section. We preserve source identifiers and utterance spans, choose interactions whose original task is unresolved, and construct a five-message Assistant, User, Assistant, User, Assistant exchange with DeepSeek-V4-Pro at temperature zero. Conversation construction receives the legitimate task but not the attacker goal, benchmark attack, payload, or outcome. The resulting interaction is frozen before the attacker sees it and does not execute tools or alter benchmark state.

\subsection{Victims, injection, and outcome accounting}
The evaluated victim set is GLM-5.1, MiniMax-M2.7, DeepSeek-V4-Flash, Qwen3.5-27B, Qwen3.6-35B-A3B, and Qwen3-235B-A22B-Instruct-2507. Target calls use temperature zero. The main paper's model comparison displays the first four victims; its policy interaction figure displays Qwen3.6-35B-A3B and DeepSeek-V4-Flash. These are different result slices and must not be read as a single pooled six-model estimate.

For each evaluated condition, the attacker receives its fixed malicious goal before read-only observation of the frozen benign interaction. It extracts source-verifiable candidate evidence, selects an item, and constructs one complete payload for the designated untrusted benchmark channel. Earlier turns, the legitimate task, and the benchmark success predicate remain outside attacker control. The evaluation record associates the goal, selected source span, policy and rewriting labels, injected message, victim output, tool trace, and benchmark outcome. An unavailable or invalid evaluation is recorded separately from a valid success or failure.

Attack success rate (ASR) is the fraction of evaluated attacks satisfying the benchmark's malicious-goal predicate. AgentDojo also reports utility under attack (UA), completion of the trusted task, and joint success rate (JSR), completion of both goals. InjecAgent reports ASR for Direct Harm and Data Stealing. In the main comparison table, each AgentDojo entry is ordered ASR, UA, JSR; InjecAgent entries contain ASR only. An Overall value belongs to the benchmark and model scope indicated by its table or figure. Percentage-point changes compare ASR values within a stated matched result block.

\section{Evidence Extraction and Attack Construction}
\label{app:method_details}

\subsection{Source-linked evidence}
For case $i$, the attacker knows $g_i^a$ while observing the permitted read-only interaction $\mathcal{I}_{i,K}^{A}$. The extractor returns candidates with a source span, one or more category labels, source role, source turn, and a pointer to the supporting words:
\begin{equation}
\mathcal{E}_{i,K}=\mathcal{O}_A(\mathcal{I}_{i,K}^{A},g_i^a),\qquad
 e_j=\langle z_j,L_j,\rho_j,t_j,\pi_j\rangle.
\end{equation}
An item is verifiable when its pointer resolves to an earlier benign turn. This verifies provenance; authorization for the requested action remains a separate question. The benign user's Goal label is distinct from the attack goal $g_i^a$.

\begin{table}[t]
\centering
\small
\begin{tabularx}{\columnwidth}{@{}p{0.25\columnwidth}X@{}}
\toprule
Stage 1 evidence category & Source-linked content \\
\midrule
Events & An occurrence, date, deadline, or other time-linked event. \\
Operational & A resource, environment condition, or operating constraint. \\
Preference & A stated priority, choice, or constraint on an outcome. \\
Relationship & A person or organization and its role in the task. \\
Task Status & Progress, a failed attempt, or a pending action. \\
Goal & The benign user's intended outcome. \\
\bottomrule
\end{tabularx}
\caption{The six Stage 1 evidence categories. A source item may carry multiple labels.}
\label{tab:evidence_taxonomy}
\end{table}

\subsection{Evidence Extracting Policy}
The selector receives the candidate set and the fixed attack goal. The main Method figure names the three policies User Oriented, Neutral, and Task Oriented. The registered result labels are User-Events, Randomized, and Task-Events, respectively. Selection keeps the source pointer attached to the chosen item. The main paper's ASR heatmap and joint effects figure use these registered labels.

\subsection{Rewriting policy and reference condition}
Stage 2 combines the goal, injection context, selected evidence when available, and assigned condition into one untrusted message:
\begin{equation}
x_{i,s}=G_{\mathrm{attack}}(g_i^a,\chi_i,\widetilde e_{i,s};s).
\end{equation}
The conditions are Low-Impact, Strategic, Time-Sensitive, and Critical. Low-Impact is the reference; Strategic frames the requested action as important through rhetorical register; Time-Sensitive quotes or faithfully paraphrases the evidence to press for prompt action; Critical asserts a consequence, responsibility, or repair need to make the action appear urgent and important. The evidence item's source support and the added evidence-to-action relation are assessed separately. The main paper's matched ASR-gain figure uses a Low-Impact reference without benign-conversation exposure, while the category comparison in the evidence-group figure includes the selected item in Low-Impact. Each figure's reference is therefore interpreted within its own matched comparison.

\subsection{Reflection record}
\label{app:reflection}
After the benchmark execution, reflection links the victim's output and official outcome to the selected evidence, extraction policy, and rewriting condition. When a failed attack elicits an explicit reason for declining the malicious goal, the record retains the victim's words and whether the objection concerns the cited evidence, an asserted relation, or the requested action. Generic refusals, failures with no stated reason, and handling of the trusted task are separate annotations. The resulting labels characterize stated objections and observed outcomes; attack success remains determined by the benchmark predicate.

\section{Detailed Evaluation Results}
\label{app:additional_results}
The supplementary tables below transcribe the current model comparison and plotted result data. The cross-benchmark method table in the main paper reports AgentDojo and InjecAgent separately, and the matched rewriting, policy-interaction, evidence-group, and defense tables each have their own result populations. In particular, the six single-label source groups in the evidence-category plot are a displayed result grouping; Stage 1 evidence records may have overlapping category labels. A model-specific value or a Best stratum is not an estimate over all six victims or all benchmark suites.

The defense plots report point estimates for four attack conditions across five defenses. Their supplied data do not include cell-level denominators or uncertainty intervals, so the appendix does not assign a significance level to their differences. The matched rewriting comparisons retain the intervals and case counts available in their plot data. All figures and tables in this appendix use the current names Evidence Extracting Policy and rewriting policy.

\subsection{Victim-model outcomes}
The main paper's model comparison figure compares six methods across four displayed victims. Table~\ref{tab:appendix_model_results} gives its ASR, UA, and JSR values in the same order as the main method table. The remaining evaluated victims appear only in the result slices named in the main text.
\begin{table*}[t]
\centering
\scriptsize
\setlength{\tabcolsep}{3pt}
\resizebox{\textwidth}{!}{%
\begin{tabular}{lcccc}
\toprule
Method & GLM-5.1 & MiniMax-M2.7 & DeepSeek-V4-Flash & Qwen3.5-27B \\
\midrule
Benchmark & 6.7 / 68.0 / 3.3 & 10.5 / 69.2 / 3.3 & 23.9 / 74.7 / 11.3 & 20.3 / 73.7 / 5.2 \\
AgentVigil & 10.3 / 83.2 / 4.2 & 16.0 / 76.0 / 3.5 & 28.8 / 65.8 / 6.9 & 21.9 / 79.8 / 5.6 \\
IterInject & 9.9 / 75.5 / 4.5 & 16.9 / 73.6 / 5.7 & 34.1 / 77.8 / 15.3 & 24.0 / 77.5 / 5.8 \\
Muzzle & 9.3 / 79.5 / 5.0 & 16.9 / 76.5 / 7.4 & 36.4 / 72.6 / 11.4 & 30.2 / 67.9 / 15.0 \\
ChatInject & 10.3 / 76.2 / 4.7 & 18.0 / 74.0 / 6.9 & 37.4 / 70.9 / 10.7 & 32.3 / 69.5 / 15.9 \\
ContextLure & 11.2 / 79.1 / 4.9 & 20.0 / 75.2 / 6.8 & 41.1 / 69.2 / 10.0 & 33.8 / 67.4 / 16.3 \\
\bottomrule
\end{tabular}%
}
\caption{Model-level method results corresponding to the main paper's model comparison figure. Each cell is ASR / UA / JSR (\%); values are model-specific.}
\label{tab:appendix_model_results}
\end{table*}
\subsection{Matched rewriting-policy changes}
Table~\ref{tab:appendix_matched_rewriting} retains the case counts, gain/loss counts, and 95\% intervals behind the main paper's matched ASR-gain figure. A gain is a case where the named condition succeeds and its Low-Impact reference fails; a loss is the reverse. The displayed change is the matched ASR difference in percentage points.
\begin{table*}[t]
\centering
\small
\resizebox{\textwidth}{!}{%
\begin{tabular}{llrrrrl}
\toprule
Benchmark and stratum & Rewriting policy & $N$ & Gain & Loss & $\Delta$ ASR (pp) & 95\% CI (pp) \\
\midrule
AgentDojo Best (Travel) & Strategic & 34 & 9 & 2 & +20.6 & [2.9, 38.2] \\
 & Time-Sensitive & 34 & 16 & 2 & +41.2 & [20.6, 61.8] \\
 & Critical & 34 & 9 & 5 & +11.8 & [-8.8, 32.4] \\
\addlinespace
AgentDojo Overall & Strategic & 74 & 14 & 15 & -1.4 & [-14.9, 13.5] \\
 & Time-Sensitive & 74 & 20 & 9 & +14.9 & [1.4, 28.4] \\
 & Critical & 74 & 15 & 14 & +1.4 & [-12.2, 16.2] \\
\addlinespace
InjecAgent Best & Strategic & 62 & 25 & 1 & +38.7 & [25.8, 51.6] \\
 & Time-Sensitive & 62 & 29 & 0 & +46.8 & [33.9, 59.7] \\
 & Critical & 62 & 29 & 0 & +46.8 & [33.9, 59.7] \\
\addlinespace
InjecAgent Overall & Strategic & 198 & 82 & 3 & +39.9 & [32.8, 47.0] \\
 & Time-Sensitive & 198 & 90 & 2 & +44.4 & [37.4, 51.5] \\
 & Critical & 198 & 93 & 1 & +46.5 & [39.4, 53.5] \\
\bottomrule
\end{tabular}%
}
\caption{Matched ASR changes plotted in the main paper's matched ASR-gain figure. Best identifies Travel for AgentDojo and the Representative cohort for InjecAgent; Overall identifies the Extension cohort in each benchmark. Each comparison uses the Low-Impact reference in the same result block.}
\label{tab:appendix_matched_rewriting}
\end{table*}
\subsection{Evidence Extracting Policy and rewriting}
Table~\ref{tab:appendix_policy_matrix} gives the absolute ASR values underlying the main paper's policy interaction figure. Neutral, User Oriented, and Task Oriented are the names used in the Method figure; Randomized, User-Events, and Task-Events are their registered result labels.
\begin{table}[t]
\centering
\small
\resizebox{\columnwidth}{!}{%
\begin{tabular}{llrrrr}
\toprule
Victim & Evidence Extracting Policy & Low-Impact & Strategic & Time-Sensitive & Critical \\
\midrule
Qwen3.6-35B-A3B & Randomized & 16.8 & 27.6 & 42.4 & 42.4 \\
 & User-Events & 24.4 & 38.1 & 37.5 & 50.4 \\
 & Task-Events & 16.8 & 42.4 & 84.0 & 42.4 \\
\addlinespace
DeepSeek-V4-Flash & Randomized & 18.9 & 49.5 & 37.9 & 43.2 \\
 & User-Events & 21.1 & 46.3 & 42.1 & 44.2 \\
 & Task-Events & 15.8 & 45.3 & 44.2 & 46.3 \\
\bottomrule
\end{tabular}%
}
\caption{ASR (\%) by victim, Evidence Extracting Policy, and rewriting policy for the two victims in the main paper's policy interaction figure. Low-Impact is the within-policy reference for its plotted percentage-point changes.}
\label{tab:appendix_policy_matrix}
\end{table}
The main paper's evidence-group figure uses the six displayed single-label source groups in Table~\ref{tab:appendix_evidence_groups}. Its rows are shown under the closest Stage 1 category names for readability; these source groups should not be counted as the complete, potentially overlapping Stage 1 label distribution.
\begin{table}[t]
\centering
\small
\resizebox{\columnwidth}{!}{%
\begin{tabular}{lrrrrr}
\toprule
Displayed source group & $N$ & Low-Impact & Strategic & Time-Sensitive & Critical \\
\midrule
Events & 43 & 11.6 & 46.5 & 41.9 & 39.5 \\
Goal & 138 & 14.5 & 45.7 & 40.6 & 42.8 \\
Relationship & 15 & 40.0 & 73.3 & 53.3 & 60.0 \\
Operational & 21 & 38.1 & 52.4 & 47.6 & 47.6 \\
Preference & 16 & 12.5 & 56.2 & 43.8 & 50.0 \\
Task Status & 39 & 28.2 & 43.6 & 46.2 & 56.4 \\
\bottomrule
\end{tabular}%
}
\caption{DeepSeek-V4-Flash ASR (\%) for the displayed single-label source groups in the main paper's evidence-group figure. $N$ counts records in each source group; Low-Impact includes the selected evidence item in this comparison.}
\label{tab:appendix_evidence_groups}
\end{table}
\subsection{Defense outcomes}
Table~\ref{tab:appendix_defense_outcomes} reports both panels of the main paper's defense figure. The combined Evidence + Rewriting condition has the largest ASR without a defense, 44.6\%; PI detector reduces that condition to 6.2\% ASR and reports 28.2\% UA. These are point estimates from the displayed defense data.
\begin{table*}[t]
\centering
\small
\resizebox{\textwidth}{!}{%
\begin{tabular}{lcccc}
\toprule
Defense & Default & Without Rewriting & Without Evidence & Evidence + Rewriting \\
\midrule
No defense & 24.6 / 24.7 & 29.5 / 26.0 & 38.0 / 14.0 & 44.6 / 18.5 \\
PI detector & 1.9 / 36.0 & 2.1 / 23.0 & 4.6 / 16.3 & 6.2 / 28.2 \\
Lakera Guard & 3.1 / 35.0 & 8.0 / 24.6 & 18.9 / 24.9 & 16.5 / 19.6 \\
Instructional prevention & 5.8 / 18.0 & 13.9 / 32.0 & 12.8 / 21.0 & 19.4 / 25.0 \\
Data delimiting & 5.2 / 19.0 & 11.6 / 22.0 & 22.1 / 20.0 & 27.6 / 23.0 \\
User-instruction repetition & 2.3 / 17.0 & 13.3 / 14.0 & 23.6 / 15.0 & 21.6 / 26.0 \\
\bottomrule
\end{tabular}%
}
\caption{AgentDojo defense outcomes corresponding to the main paper's defense figure. Each cell is ASR / UA (\%). The available plotted data do not provide cell-level denominators or confidence intervals.}
\label{tab:appendix_defense_outcomes}
\end{table*}

\section{Mitigation and Defense Details}
\label{app:mitigation}

\subsection{Evaluated defense conditions}
The main defense figure compares PI detector and Lakera Guard with instructional prevention, data delimiting, and user-instruction repetition on AgentDojo. It crosses each defense with Default, Without Rewriting, Without Evidence, and Evidence + Rewriting attack conditions. Table~\ref{tab:appendix_defense_outcomes} retains the ASR and UA values for every displayed cell. These comparisons keep the attack condition visible so that a detector's effect is not inferred from one attack form alone.

All five defenses lower ASR relative to the corresponding undefended condition. For Evidence + Rewriting, ASR is 44.6\% without a defense, 6.2\% with PI detector, 16.5\% with Lakera Guard, 19.4\% with instructional prevention, 27.6\% with data delimiting, and 21.6\% with user-instruction repetition. The same cells report UA of 18.5\%, 28.2\%, 19.6\%, 25.0\%, 23.0\%, and 26.0\%, respectively. The available figure data report point estimates, with no cell-level denominators or confidence intervals; the observed ordering is descriptive.

\subsection{Authorization at the action boundary}
The attack uses an earlier benign turn as a premise for a later unauthorized action. Verifying that a cited event, preference, or failed task occurred does not grant permission for the requested transfer, disclosure, or write. A practical control should retain the source and trust status of recalled evidence, compare proposed tool arguments with the trusted user's task and an independently maintained permission record, and request authenticated confirmation for a high-impact action when that grant is insufficient. These are application controls proposed from the observed failure mode; they are distinct from the five evaluated defense conditions.

For example, a source-linked note that a transfer failed can inform the agent's explanation of the failure. It does not authorize changing the beneficiary. The decision must be made against the trusted task and the proposed tool arguments at the time of execution. Rechecking each externally visible or state-changing call also limits propagation from an apparently helpful intermediate step to an unauthorized follow-up action. Applications can record the grant, relevant source span, decision, and executed arguments for audit while limiting retention of private conversation content.

The defense comparisons show that the evaluated detectors and prompt-based controls reduce but do not eliminate success under Evidence + Rewriting. Testing the proposed authorization controls would require matched cases with those controls implemented and a separate measurement of benign task completion and false blocking. We do not infer the effectiveness of an untested authorization rule from the existing defense bars.

\end{document}